\documentclass[10pt,twocolumn,letterpaper]{article}

\usepackage{wacv}              

\usepackage{graphicx}
\usepackage{amsmath}
\usepackage{amssymb}
\usepackage{booktabs}

\usepackage{times}
\usepackage{epsfig}

\newcommand{\bg}[1]{\boldsymbol{#1}} 
\newcommand{\bm}[1]{\mathbf{#1}} 

\newcommand\T{{\mathpalette\raiseT\intercal}}
\newcommand\raiseT[2]{%
\setbox0\hbox{$#1{#2}$}\raise\dp0\box0}

\usepackage{booktabs}
\usepackage{subcaption}
\usepackage{makecell}
\usepackage{pythonhighlight}

\usepackage[ruled,vlined]{algorithm2e}
\definecolor{commentcolor}{RGB}{24, 158, 159}   

\usepackage[pagebackref,breaklinks,colorlinks]{hyperref}

\usepackage[capitalize]{cleveref}
\crefname{section}{Sec.}{Secs.}
\Crefname{section}{Section}{Sections}
\Crefname{table}{Table}{Tables}
\crefname{table}{Tab.}{Tabs.}

\def\wacvPaperID{} 
\def\confName{WACV}
\def\confYear{2024}

\begin{document}

\title{Learning to Recognize Occluded and Small Objects with Partial Inputs}

\author{Hasib Zunair\\
CIISE, Concordia University\\
Montreal, QC, Canada\\
{\tt\small hasibzunair@gmail.com}
\and
A. Ben Hamza\\
CIISE, Concordia University\\
Montreal, QC, Canada\\
{\tt\small hamza@ciise.concordia.ca}
}
\maketitle

\begin{abstract}
Recognizing multiple objects in an image is challenging due to occlusions, and becomes even more so when the objects are small. While promising, existing multi-label image recognition models do not explicitly learn context-based representations, and hence struggle to correctly recognize small and occluded objects. Intuitively, recognizing occluded objects requires knowledge of partial input, and hence context. Motivated by this intuition, we propose Masked Supervised Learning (MSL), a single-stage, model-agnostic learning paradigm for multi-label image recognition. The key idea is to learn context-based representations using a masked branch and to model label co-occurrence using label consistency. Experimental results demonstrate the simplicity, applicability and more importantly the competitive performance of MSL against previous state-of-the-art methods on standard multi-label image recognition benchmarks. In addition, we show that MSL is robust to random masking and demonstrate its effectiveness in recognizing non-masked objects. Code and pretrained models are available on \href{https://github.com/hasibzunair/msl-recognition}{GitHub}.
\end{abstract}
  \section{Introduction}
  Multi-label image recognition (MLIR) is a fundamental and challenging task in a variety of computer vision applications such as automatic tagging of images on social media platforms and object detection in autonomous vehicles~\cite{LongChen:19}. The aim is to recognize multiple objects or attributes in an image. A major challenge in MLIR is how to effectively tackle the issue of large variations in the size and spatial locations of objects. This issue becomes more pronounced when the objects are occluded and small.

  Recent MLIR approaches, including graph convolutional networks and their variants~\cite{chen2019multi, chen2021learning,ye2020attention}, focus primarily on capturing semantics and label co-occurrence among objects. While powerful, most of these methods require the combination of multiple networks, resulting in high computation cost. Also, methods that deal with both semantics of objects and label relations often consist of multiple stages of training~\cite{liu2021query2label,liu2018multi}, rely on large language models~\cite{li2020bi}, and operate on high input resolution~\cite{chen2019learning,ye2020attention, gao2021learning,lanchantin2021general}. Moreover, they require additional data for pretraining~\cite{chen2020knowledge}, even with models already pretrained on large datasets such as ImageNet-1k and ImageNet-21k, and also rely on complex data augmentation strategies~\cite{ben2020asymmetric,zhu2021residual}. In addition, these methods do not explicitly address the occlusion problem and fail to accurately recognize small objects, leading to suboptimal performance on images containing small and occluded objects. In practical real-world applications of MLIR such as object detection in self-driving cars, images are usually comprised of multiple objects of different sizes (e.g., small) and shapes that co-exist and are densely cluttered (e.g., occluded), and hence it is of vital importance to develop MLIR approaches that can effectively recognize small objects even under heavy occlusions.

  Intuitively, we can consider occluded objects as \textit{partial inputs}, and hence accurate recognition requires knowledge of \textit{partial inputs}, and hence context. Motivated by this intuition, we propose \textit{Masked Supervised Learning (MSL)}, a single-stage, model-agnostic learning paradigm for MLIR tasks. Given a base recognition network, MSL uses a masked branch to predict the labels for a heavily masked version of the input image, which is a good cue for learning context-based representations. We also propose to use label consistency to model label co-occurrence by maximizing the similarity between the predictions from the recognition and masked branches. The main contributions of this work can be summarized as follows:
  \begin{itemize}
  \item We propose a simple yet effective single-stage, model-agnostic learning paradigm that aims to learn context-based representations and to better model label co-occurrence from \textit{partial inputs} via masking.
  \item We demonstrate through experimental results and ablations that MSL yields competitive performance in comparison with single- and multi-stage approaches, especially for small and occluded objects.
  \item We show that MSL is not only robust to \textit{partial inputs}, but also predicts objects that are almost entirely masked, while yielding improved recognition of non-masked objects.
  \end{itemize}

  \section{Related Work}
  \noindent\textbf{Hybrid Methods.}\quad These methods leverage a combination of convolutional, graph, transformer or recurrent neural
  networks~\cite{zhu2021residual,chen2019multi,chen2018recurrent,lanchantin2021general,chen2022sst}. Graph based networks, for instance, leverage semantic relations between object classes~\cite{chen2019multi, chen2021learning}, but tend to incur heavy computation costs. ADD-GCN~\cite{ye2020attention} dynamically generates graphs for an image by first generating a category-aware representation, followed by modeling the relationship between the representations. ADD-GCN~\cite{ye2020attention} operates on high resolution in the same vein as SSGRL~\cite{chen2019learning}, C-Tran~\cite{lanchantin2021general} and MCAR~\cite{gao2021learning}. KGGR~\cite{chen2020knowledge} operates on knowledge graphs and requires additional data for pretraining. Our method does not require the combination multiple networks, high input resolution, or additional data.

  \medskip \noindent\textbf{Model-Agnostic Methods.}\quad This class of approaches are not architecture dependent, and include ASL~\cite{ben2020asymmetric} and CSRA~\cite{zhu2021residual}, which can be applied to any architecture, but require an exhaustive hyperparameter tuning. Moreover, they achieve competitive results only when using complex data augmentation techniques such as CutMix, GPU Augmentations, or RandAugment~\cite{ben2020asymmetric,zhu2021residual}. By comparison, our proposed model achieves state-of-the-art performance without relying on complex data augmentation strategies.

  \medskip \noindent\textbf{Multistage and Bimodal Frameworks.}\quad Query2Label (Q2L)~\cite{liu2021query2label} is a two-stage framework that focuses on class-specific attention. KSS-Net~\cite{liu2018multi} is a knowledge distillation based method comprised of a two-stage training scheme with teacher and student models. BMML~\cite{li2020bi} is a bimodal learning approach that not only uses a convolutional neural network and a recurrent neural network, but also relies on large language models~\cite{devlin2018bert} and additional data. Our work differs from these frameworks in that it does not require multiple stages of training and also does not rely on large language models.

  \medskip\noindent\textbf{Transformer-Based Methods.}\quad TDRG~\cite{zhao2021transformer} consists of convolutional neural network, a transformer, as well as a graph neural network that is used to capture long-term contextual information and to build position-wise relationships at different scales. C-Tran~\cite{lanchantin2021general} is a transformer based method that relies on an additional image feature extractor and high input resolution. It exploits the dependencies among both visual features and labels using a single transformer encoder. By comparison, our work is significantly different from C-Tran. First, during training we mask images, whereas C-Tran masks labels. Second, the input to the transformer encoder in C-Tran consists of an image and a masked label (i.e., token), whereas our model requires only an image. Also, our method is model-agnostic and can be applied to any kind of network for MLIR tasks.

  \medskip
  Overall, our work differs from previous MLIR approaches in that we propose a simple yet effective single-stage learning paradigm that is model-agnostic. Most notably, our model does not require multiple stages of training, the combination of multiple networks, large language models, high input resolution, complex data augmentation strategies, or additional data for pretraining.

\section{Masked Supervised Learning}
\begin{figure*}[!htp]
  \centering
  \includegraphics[scale=1.0]{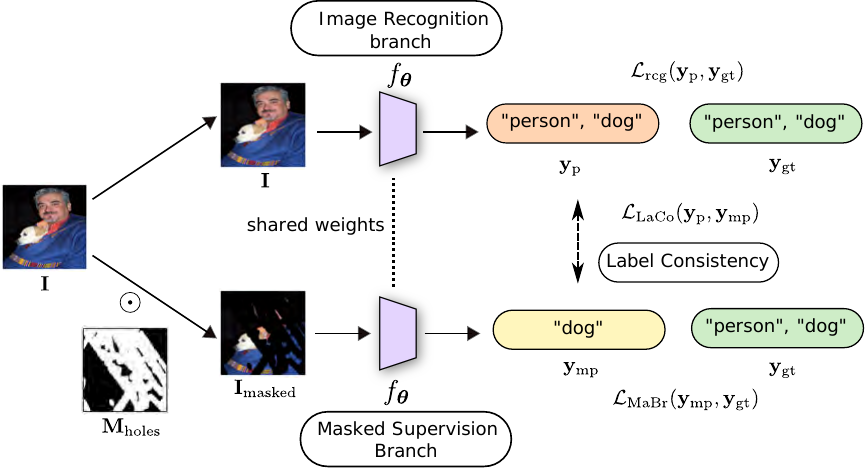}
  \caption{\textbf{Overview of Masked Supervised Learning (MSL)}. A single-stage model-agnostic training scheme with a Masked Branch (MaBR) and Label Consistency (LaCo) for MLIR tasks where $f_{\bg{\theta}}$ is a base network. The recognition and masked branches are identical and share weights. After training, $f_{\bg{\theta}}$ is used to obtain the multi-label prediction.}
  \label{Fig:schema}
\end{figure*}

  In this section, we begin by formulating the task at hand and subsequently introduce the fundamental components that make up the proposed MSL paradigm. The overall framework of MSL is depicted in Figure~\ref{Fig:schema}.

\medskip\noindent\textbf{Problem Statement.}\quad Let $\mathcal{D}=\{(\bm{I}_i,\bm{y}_i)\}_{i=1}^{N}$ be a training set of $N$ labeled images $\bm{I}_{i}\in\mathcal{X}$ and their ground-truth multi-label vectors $\bm{y}_{i}=(y_{i,1},\dots,y_{i,K})^{\T}\in\mathcal{Y}=\{0,1\}^{K}$, with $y_{i,k}=1$ indicating the presence of the $k$-th label (i.e., object or attribute) in the image, and $y_{i,k}=0$ indicating its absence. In other words, each image $\bm{I}_{i}$ is associated with multiple labels chosen from a set of $K$ possible classes (i.e., object categories). The task of multi-label image recognition is to learn a multi-label recognition model $f_{\bg{\theta}}: \mathcal{X}\to \mathcal{Y}$, where $\bg{\theta}$ is a set of learnable parameters. Given a test image $\bm{I}$, the trained model predicts the corresponding multi-label vector $\bm{y}_\text{p}=\sigma(f_{\bg{\theta}}(\bm{I}))$, where $\sigma(\cdot)$ is the sigmoid activation function applied element-wise.

\subsection{Masked Inputs}
For masked image generation, we leverage the Irregular Mask dataset~\cite{liu2018image}, which is commonly used in image inpainting~\cite{liu2018image,suvorov2022resolution} and is comprised of roughly $20,000$ masks with random streaks and holes of arbitrary shapes. From this dataset, we generate \textit{low} and \textit{high} mask subsets, each of which is comprised of 1000 samples. The process for creating these two subsets is as follows: For a given mask sampled from the Irregular Mask dataset, we first compute the percentage $p$ of zero pixels in the mask. If $p$ is greater than 50\%, then the mask is included in the high mask subset. Otherwise, the mask is placed in the low mask subset. In our experiments, we find that \textit{high} masks generally improve performance. Intuitively, image masking can be viewed as ``simulating" images with partial inputs. During training, we randomly sample a mask from the \textit{high} mask subset and perform binary thresholding, where the pixel values are either 0 or 1, and we denote this mask by $\bm{M}_\text{holes}$. Then, we follow the masking procedure in~\cite{liu2018image,Zunair2022MSL} to create a masked image $\bm{I}_\text{masked} = \bm{I}\odot \bm{M}_\text{holes}$, where $\bm{I}$ is the input image, and $\odot$ denotes element-wise multiplication. The masked image has a similar layout as the input image, but with roughly 50\% of pixels randomly removed.

  \subsection{Masked Branch}
  The goal of Masked Branch (MaBr) is to explicitly learn context-based representations, as this branch is tasked to predict labels of heavily masked inputs (i.e., \textit{partial inputs}), translating into better multi-label predictions. \textbf{The masked branch has the ability to learn short-range context even when objects in the image are densely cluttered. Similarly, it can also learn long-range context when objects are more spaced apart}.

  Given an input image $\bm{I}$ and its masked version $\bm{I}_\text{masked}$, we train a base recognition network $f_{\bg{\theta}}$ to predict both the output $\bm{y}_\text{p}$ of the image recognition branch and the output $\bm{y}_\text{mp}$ of the masked branch. Here, $f_{\bg{\theta}}$ is a Siamese network like architecture~\cite{bromley1993signature}, where the branches are identical and share weights.

  We train $f_{\bg{\theta}}$ by minimizing the following combined loss function of the recognition branch and masked branch
  \begin{equation}
  \mathcal{L}_\text{inter} = \mathcal{L}_\text{rcg} (\bm{y}_\text{p}, \bm{y}_\text{gt}) + \mathcal{L}_\text{MaBr} (\bm{y}_\text{mp}, \bm{y}_\text{gt}),
  \label{eq:loss1}
  \end{equation}
  where $\mathcal{L}_\text{rcg}$ and $\mathcal{L}_\text{MaB}$ are binary cross-entropy losses between the ground truth and the outputs of the recognition and masked branch, respectively. Application-specific loss functions can also be used in lieu of cross-entropy.

  \subsection{Label Consistency}
  As objects generally co-exist in an image (e.g., \textit{chair} is more likely to co-occur with \textit{table} than a \textit{sportsball}), it is of vital importance to model this label co-occurrence to help improve the recognition performance. \textbf{To perceive this label-level feature, we propose to use Label Consistency (LaCo) that maximizes the similarity between the predictions from the recognition and masked branch}. Since we use a Siamese style architecture, where the network is the same with shared weights, maximizing the predictions helps the network learn to predict heavily occluded objects (e.g., partial inputs) from the presence of other target objects, thereby effectively utilizing masked branch. More specifically, we maximize the similarity between the predictions from the recognition branch $\bm{y}_\text{p}$ and the masked branch $\bm{y}_\text{mp}$ by minimizing the $L_2$-loss $\mathcal{L}_\text{LaCo}=\Vert\bm{y}_\text{p}-\bm{y}_\text{mp}\Vert^{2}$.

  \subsection{Overall Loss Function}
  Using the recognition branch, masked branch and label consistency, we define the overall loss function for the proposed MSL model as follows
  \begin{equation}
  \begin{aligned}
  \mathcal{L}_\text{total} & = \alpha_{1} \mathcal{L}_\text{rcg} (\bm{y}_\text{p}, \bm{y}_\text{gt}) + \alpha_{2} \mathcal{L}_\text{MaBr} (\bm{y}_\text{mp}, \bm{y}_\text{gt}) \\
  &\quad + \alpha_{3} \mathcal{L}_\text{LaCo} (\bm{y}_\text{p}, \bm{y}_\text{mp}),
     \end{aligned}
  \label{eq:loss2}
  \end{equation}
where the scalars $\alpha_{1}$, $\alpha_{2}$ and $\alpha_{3}$ are nonnegative trade-off hyperparameters, which control the contribution of each loss term.

During training, $\mathcal{L}_\text{total}$ is minimized between predictions and ground-truth labels for several epochs using stochastic gradient descent to learn the parameters of $f_{\bg{\theta}}$ using a labeled training set. For inference, the trained network $f_{\bg{\theta}}$ is used in multi-label image recognition to obtain multi-label predictions given a test image $\bm{I}$. Hence, MSL is simple in structure (i.e., model-agnostic) and easy to implement (i.e., single-stage training). When $\alpha_1=1$ and $\alpha_2=\alpha_3=0$, we obtain the recognition loss, which is basically the loss function for the vanilla network.


  \section{Experiments}
  In this section, we demonstrate the performance of MSL in comparison with state-of-the-art methods. Details on the implementation,
  architecture and training, as well as additional results are included in the supplementary material.

  \subsection{Experimental Setup}
  \noindent\textbf{Datasets.}\quad We conduct experiments on two MLIR benchmarks: VOC2007~\cite{everingham2010pascal} and MS-COCO~\cite{lin2014microsoft}.

  \begin{itemize}
  \item \textbf{VOC2007.} This is a widely-used dataset for MLIR tasks, and is comprised of 9,963 images with 20 classes, where the \textit{train-val} set
  has 5,011 images and the \textit{test} set has 4,952 images. Following previous work~\cite{zhu2021residual,chen2022sst}, we use the \textit{train-val} for training and \textit{test} for testing. We also set the input resolution to $448\times 448$, unless otherwise specified.
  \item \textbf{MS-COCO.}\quad This is a standard benchmark for training and evaluating image recognition, segmentation, and detection algorithms. In our experiments, we use COCO-2014, which consists of 82,081 and 40,137 training and validation images, respectively, with 80 different classes. For fair comparison with previous work~\cite{zhu2021residual,zhao2021transformer,lanchantin2021general}, we use the same training and evaluation procedures, and evaluation metrics.
  \end{itemize}

  \noindent\textbf{Baselines.}\quad We compare MSL against several state-of-the-art graph-based methods that use different learnable networks such as ML-GCN~\cite{chen2019multi}, P-GCN~\cite{chen2021learning}, ADD-GCN~\cite{ye2020attention} and TDRG~\cite{zhao2021transformer}. We also compare against model-agnostic methods such as ASL~\cite{ben2020asymmetric} and CSRA~\cite{zhu2021residual}, which rely on complex data augmentation. Moreover, we compare against methods that require large language models and additional data for pretraining such as BMML~\cite{li2020bi} and KGGR~\cite{chen2020knowledge}, as well as methods that operate on high input resolution such as SSGRL~\cite{chen2019learning}, C-Tran~\cite{lanchantin2021general}, MCAR~\cite{gao2021learning}, and IDA~\cite{liu2022causality}. Finally, we compare against multi-stage frameworks such as KSS-Net~\cite{liu2018multi} and Query2Label~\cite{liu2021query2label}.

  \medskip\noindent\textbf{Evaluation Metrics.}\quad We use the mean average precision (mAP) as primary evaluation metric~\cite{zhu2021residual,chen2022sst}. We set positive threshold to 0.5 and report overall performance results of MSL and baselines using other evaluation metrics, including overall precision (OP), overall recall (OR), overall F1-measure (OF1), per-category precision (CP), per-category recall (CR), and per-category F1-measure (CF1).

  \subsection{Comparison with State-Of-The-Art}
  \noindent\textbf{Comparisons on VOC2007.}\quad We compare the performance of MSL against several state-of-the-art methods, and the results are reported in Table~\ref{Tab:voc_sota}. All scores are averaged over 3 runs. We employ MSL with two CSRA-based backbones: ResNet-cut, which is a ResNet-101~\cite{he2016deep} pretrained on ImageNet-1k with CutMix~\cite{yun2019cutmix}, and ViT-L16~\cite{dosovitskiy2020vit}, which is a large vision Transformer pretrained on ImageNet-1k with $224 \times 224$ resolution. We refer to these MSL variants as MSL-C and MSL-V, respectively. The classification head of these backbones differs from the typical fully connected or global average pooling layer by utilizing a CSRA module~\cite{zhu2021residual}. This module generates class-specific features for each category, and then combines the intermediate results to produce the final logits. As shown in the table, MSL-C outperforms all previous state-of-the-art models, achieving relative improvements of 1.1\%, 5.6\% and 3.9\% in terms of mAP, CR and CF1, respectively, over the strongest baseline. MSL-C performs better than graph-based methods such as ML-GCN and ADD-GCN. MSL-C also achieves a relative improvement of 2.8\% in terms of mAP over SSGRL, which is trained on input resolution of $640\times 640$ and uses both a convolutional feature extractor and a graph neural network. Notably, MSL-C is also efficient and more accurate than KGGR and BMML, which use additional data (MS-COCO) consisting of 82,081 images for pretraining on top of ImageNet-1k pretraining, and also rely on large language model BERT~\cite{devlin2018bert} and operate on label-level attentions (i.e, multiple images), making them compute intensive.

\begin{figure*}[htp]
  \centering
  \includegraphics[scale=.195]{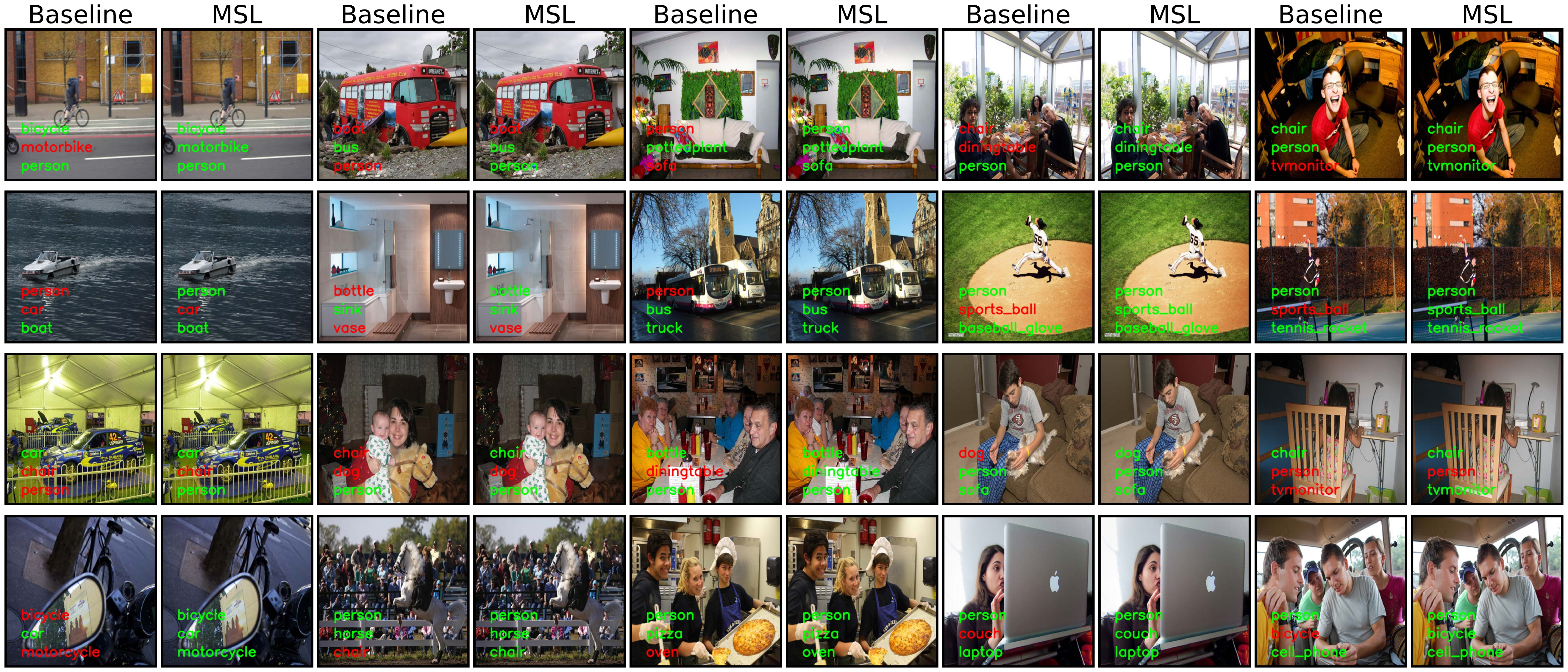}
  \caption{\textbf{Visual comparison of MSL and CSRA~\cite{zhu2021residual} on VOC2007 and MS-COCO}. First two rows show samples from VOC2007: the first row shows cases where MSL can better recognize small objects and the second row shows cases of heavy occlusions. The last two rows shows samples from MS-COCO. For both datasets, MSL is effective at recognizing small and occluded objects compared to the CSRA baseline. Zoom-in for better details.}
  \label{Fig:qual1}
  \end{figure*}

  The first two rows of Figure~\ref{Fig:qual1} show visual examples of predictions made by MSL-C and CSRA ResNet-cut as baseline. In the first row, we can see that the baseline fails to recognize small objects such as \textit{motorbike}, \textit{person}, \textit{chair} and \textit{tvmonitor}. The second row shows instances where the baseline model fails to recognize target objects under heavy occlusions, such as \textit{sports ball}, \textit{person} and \textit{vase}. In contrast, MSL-C is able to recognize small objects, as well as objects that are heavily occluded. The masked branch, which is responsible for recognizing target object(s) from partial inputs through masking, can acquire context-based representations. This ability is likely responsible for its success in recognizing objects under challenging conditions. Label consistency, on the other hand, helps model label co-occurrence by maximizing the similarity between the predictions made by the recognition and masked branches.

  \begin{table}[htp]
     \begin{center}
     \caption{\textbf{Performance comparison of MSL and baselines on VOC2007 using mAP, CR and CF1 metrics}. Boldface numbers
     indicate the best performance, whereas the best baselines are underlined. $\dagger$ indicates the results reproduced by the
     corresponding released codes or their modified versions. ``\textit{pre}'' means pretrained on the MS-COCO dataset.}
     \smallskip
     \label{Tab:voc_sota}
     \begin{tabular}{l c c c}
     \toprule[1pt]
     Method & mAP & CR & CF1\\
     \toprule[1pt]
     ResNet~\cite{he2016deep} & 92.9 & - & -\\
     FeV+LV~\cite{yang2016exploit} & 92.0 & - & -\\
     Atten-Reinforce~\cite{chen2018recurrent} & 92.0 & - & -\\
     RCP~\cite{wang2017multi} & 92.5 & - & -\\
     SSGRL~\cite{chen2019learning} & 93.4 & - & -\\
     SSGRL (\textit{pre})~\cite{chen2019learning} & 95.0 & - & -\\
     ML-GCN~\cite{chen2019multi} & 94.0 & - & -\\
     ADD-GCN~\cite{ye2020attention} & 93.6 & - & -\\
     BMML$\dagger$ (\textit{pre})~\cite{li2020bi} & \underline{95.0} & - & -\\
     IDA-R101~\cite{liu2022causality}  & 94.3 & - & -\\
     ASL~\cite{ben2020asymmetric} & 94.6 & - & -\\
     MCAR~\cite{gao2021learning} & 94.8 & - & -\\
     CSRA$\dagger$~\cite{zhu2021residual} & 93.7 & \underline{87.5} & \underline{88.3}\\
     KGGR~\cite{chen2020knowledge} & 93.6 & - & -\\ 
     KGGR (\textit{pre})~\cite{chen2020knowledge} & 95.0 & - & -\\ 
     SST~\cite{chen2022sst} & 94.5 & - & -\\
     \midrule[.8pt]
     MSL-V & 95.0 & 84.8 & 89.5\\
     MSL-C & \textbf{96.1} & \textbf{92.4} & \textbf{91.6}\\
     \bottomrule[1pt]
     \end{tabular}
     \end{center}
  \end{table}

  \medskip\noindent\textbf{Comparisons on MS-COCO.}\quad In Table~\ref{Tab:coco_sota}, we report results on MS-COCO, where all scores are averaged over 3 runs and MSL is applied on CSRA-based ResNet-cut backbone. As can be seen, MSL-C outperforms all baselines operating on input resolution $448 \times 448$ by 1.4\% in terms of mAP. MSL-C also outperforms complicated and time-consuming methods such as KSSNet and MCAR, as well as methods that operate on higher input resolution $575\times 576$ such as ADD-GCN, SSGRL, and C-Tran. In particular, MSL-C outperforms MCAR by a relative improvement of 2.2\% in terms of mAP. MCAR has two network streams that are trained jointly, and at inference predictions are fused from the two streams to generate a final prediction, whereas MSL has two streams with same weights in a Siamese-style network, which is much easier to optimize, and at inference a single network is used to make predictions. Moreover, MSL-C outperforms ADD-GCN, which uses a CNN and a GCN, by a relative improvement of 1.4\% in terms of mAP.

  In the last two rows of Figure~\ref{Fig:qual1}, we show visual examples of predictions made by MSL-C and CSRA ResNet-cut as baseline on MS-COCO. A similar pattern can be observed, where MSL can recognize small objects and also objects under heavy occlusions compared to the baseline. It is worth mentioning that
  the variation of objects and their shapes or sizes are more complex in MS-COCO than those in VOC2007.

  Overall, MSL is able to \textbf{learn context-based representations and to better model label co-occurrence} by masked branch and label consistency, thereby translating to better predictions in comparison with the baselines. MSL can \textbf{better recognize small objects and also objects under heavy occlusions}. MSL is also \textbf{very simple and much easier to train}, as \textbf{it does not require multiple stages of training, the combination of multiple learnable networks, large language models, high input resolution, complex data augmentation strategies, or additional data}.

  \begin{table*}
  \begin{center}
  \caption{\textbf{Performance comparison of MSL and baselines on MS-COCO in terms of mAP and other evaluation metrics}. Boldface numbers indicate the best performance, whereas the best baselines are underlined. $\dagger$ indicates the results reproduced by the corresponding released codes or their modified versions.}
     \smallskip
     \label{Tab:coco_sota}
     \begin{tabular}{l c c c c c c c c}
     \toprule[1pt]
     Method & Input Resolution & mAP & CP & CR & CF1 & OP & OR & OF1\\
     \midrule[1pt]
     ResNet~\cite{he2016deep} & $448 \times 448$ & 79.4 & 83.4 & 66.6 & 74.0 & 86.8 & 71.1 & 78.2\\
     PLA~\cite{yazici2020orderless} & $228 \times 228$ & - & 80.4 & 68.9 & 74.2 & 81.5 & 73.3 & 77.2\\
     ResNet-cut$\dagger$~\cite{he2016deep} & $448 \times 448$ & 82.1 & 86.2 & 68.7 & 76.4 & 88.9 & 73.1 & 80.3\\
     ML-GCN~\cite{chen2019multi} & $448 \times 448$ & 83.0 & 85.1 & 72.0 & 78.0 & 85.8 & 75.4 & 80.3\\
     MS-CMA~\cite{you2020cross} & $448 \times 448$ & 83.8 & 82.9 & 74.4 & 78.4 & 84.4 & \underline{77.9} & 81.0\\
     KSSNet~\cite{liu2018multi} & $448 \times 448$ & 83.7 & 84.6 & 73.2 & 77.2 & 87.8 & 76.2 & 81.5\\
     MCAR~\cite{gao2021learning} & $448 \times 448$ & 83.8 & 85.0 & 72.1 & 78.0 & 88.0 & 73.9 & 80.3\\
     TDRG$\dagger$~\cite{zhao2021transformer}& $448 \times 448$ & 84.6 & 86.0 & 73.1 & 79.0 & 86.6 & 76.4 & 81.2\\
     CSRA$\dagger$~\cite{zhu2021residual} & $448 \times 448$ & 84.3 & 83.5	& 74.3 & 78.6 & 85.1	& 77.2 & 81.0\\
     Q2L-R101$\dagger$~\cite{liu2021query2label}& $448 \times 448$ & 84.0 & 82.0 & 75.8 & 78.8 & 83.3 & 78.8 & 81.0\\
     IDA-R101~\cite{liu2022causality} & $448 \times 448$ & 83.8 & - & - & - & - & - & -\\
     SST$\dagger$~\cite{chen2022sst} & $448 \times 448$ & 84.2 & 86.1 & 72.1 & 78.5 & 87.2 & 75.4 & 80.8\\
     P-GCN$\dagger$~\cite{chen2021learning} & $448 \times 448$ & 83.2 & 84.9 & 72.7 & 78.3 & 85.0 & 76.4 & 80.5\\
     KGGR$\dagger$~\cite{chen2020knowledge} & $448 \times 448$ & 84.3 & 85.6 & 72.7 & 78.6 & 87.1 & 75.6 & 80.9\\
     \hline
     ADD-GCN~\cite{ye2020attention} & $576 \times 576$ & \underline{85.2} & 84.7 & 75.9 & 80.1 & 84.9 & 79.4 & \underline{82.0}\\
     SSGRL~\cite{chen2019learning} & $576 \times 576$ & 83.8 & \underline{89.9} & 68.5 & 76.8 & \underline{91.3} & 70.8 &79.7\\
     C-Tran~\cite{lanchantin2021general} & $576 \times 576$ & 85.1 & 86.3 & \underline{74.3} & \underline{79.9} & 87.7 & 76.5 & 81.7\\
     MCAR~\cite{gao2021learning} & $576 \times 576$ & 84.5 & 84.3 & 73.9 & 78.7 & 86.9 & 76.1 & 81.1\\
      \midrule[.8pt]
     MSL-C & $448 \times 448$ & \textbf{86.4}	& \textbf{90.1} & \textbf{76.3}	& \textbf{80.4} & \textbf{89.1}	& \textbf{80.0} & \textbf{82.2}\\
     \bottomrule[1pt]
     \end{tabular}
     \end{center}
  \end{table*}

  \subsection{Ablation Study}
  We analyze how each of the key components of the proposed MSL framework affects the final performance. We also perform hyperparameter sensitivity analysis.

  \medskip\noindent\textbf{Effectiveness of Masked Branch.}\quad Table~\ref{Tab:abl_modules} illustrates the benefit of using masked branch tasked to make predictions, given \textit{partial inputs} by random masking. We adopt CSRA with ResNet-cut backbone as our baseline, and evaluate performance on VOC2007. We find that the masked branch improves performance in terms of mAP and other evaluation metrics. It helps learn useful representations, especially for small and occluded objects due largely to the fact that the branch is tasked to recognize \textbf{masked objects} (i.e., \textit{partial inputs}), thereby leveraging information from neighboring objects.

  \medskip\noindent\textbf{Effectiveness of Label Consistency.}\quad As shown in Table~\ref{Tab:abl_modules}, label consistency helps improves performance in terms of mAP and other metrics. This constraint essentially guides the model to make accurate predictions on masked inputs by minimizing the distance between the predictions made by the recognition and masked branches. Basically, we push the predictions of the masked branch predictions and recognition branch predictions together and learn representations for \textit{partial inputs}. As can be seen, the best performance is achieved when combining masked branch and label consistency. Also, Table~\ref{Tab:diff_archs} shows that MSL is model-agnostic, and can also improve performance of not only classical convolutional backbones, but also modern transformer backbones.

  \begin{table}[htp]
     \begin{center}
     \caption{\textbf{Effectiveness of masked branch and label consistency on MSL performance using VOC2007}. Using both masked branch and label consistency significantly improves the baseline performance.}
     \smallskip
     \label{Tab:abl_modules}
     \begin{tabular}{l c c c c c}
     \toprule[1pt]
     Method & MaBr & LaCo & mAP & CR & CF1\\
     \midrule[1pt]
     Baseline &  &  & 93.7 & 87.5 & 88.3\\
     MSL & \checkmark &  & 94.0 & 89.1 & 88.4\\
     MSL &  & \checkmark & 94.3 & 88.1 & 88.9\\
     \midrule[.8pt]
     MSL & \checkmark & \checkmark & \textbf{96.1} & \textbf{92.4} & \textbf{91.6}\\
     \bottomrule[1pt]
     \end{tabular}
     \end{center}
  \end{table}

  \begin{table}[htp]
     \setlength{\tabcolsep}{3.5pt}
    \centering
    \caption{\textbf{Comparison of different architectures trained using MSL on VOC2007 and MS-COCO}. MSL is model-agnostic and improves performance of different architectures.}
  \smallskip
     \label{Tab:diff_archs}
     \resizebox{0.45\textwidth}{!}{%
     \begin{tabular}{l c c}
     \toprule[1pt]
     Architecture & VOC2007, mAP (\%) & MS-COCO, mAP (\%)\\
     \toprule[1pt]
     ViT & 94.4 & 76.8 \\
     + MSL & \textbf{95.0} & \textbf{79.0} \\
     \midrule[.8pt]
     ResNet & 93.7 & 84.3 \\
     + MSL & \textbf{96.1} & \textbf{86.4} \\
     \bottomrule[1pt]
     \end{tabular}
     }
  \end{table}

  \medskip\noindent\textbf{Effectiveness of Binarization.}\quad Table~\ref{Tab:binarization} shows the benefit of using binarization of masking during training. We find that applying binary thresholding to the masks significantly improves performance of the baseline in terms of all metrics. This is attributed to the fact binarization yields true masking, dropping certain pixels while retaining the rest, thereby resulting in better numerical stability. Without binarization, the image is slightly offset in the pixel space when multiplied by 0.884 instead of 1, yielding a different representation in the feature space that degrades performance.

  \begin{table}[htp]
     \begin{center}
     \caption{\textbf{Ablation analysis of binarization in MSL using VOC2007}. Binarization helps achieve better numerical stability.}
     \smallskip
     \label{Tab:binarization}
     \begin{tabular}{l c c c}
     \toprule[1pt]
     Binarization & mAP & CR & CF1\\
     \midrule[1pt]
     Baseline & 93.7 & 87.5 & 88.3\\
     + w/o Binarization & 93.8 & 88.2 & 88.2\\
     + w/ Binarization & \textbf{96.1} & \textbf{92.4} & \textbf{91.6}\\
     \bottomrule[1pt]
     \end{tabular}
     \end{center}
  \end{table}

  \medskip\noindent\textbf{Amount of Masking.}\quad In Table~\ref{Tab:abl_masking}, we report the effect of the amount of masking on MSL performance. We adopt CSRA with ResNet-cut backbones, and evaluate performance on VOC2007 and MS-COCO, respectively. We find that applying extensive image masking during the training process leads to improved performance.

  \begin{table}[htp]
     \setlength{\tabcolsep}{3.5pt}
    \centering
    \caption{\textbf{Ablation analysis of \textit{high-} and \textit{low-}masked pixels during MSL training using VOC2007 and MS-COCO}. MSL with \textit{high-}masked pixels yields better performance.}
     \smallskip
     \label{Tab:abl_masking}
     \resizebox{0.45\textwidth}{!}{%
     \begin{tabular}{l c c}
     \toprule[1pt]
     Masking & VOC2007, mAP (\%) & MS-COCO, mAP (\%)\\
     \midrule[1pt]
     Low & 95.0 & 85.1 \\
     High & \textbf{96.1} & \textbf{86.4}\\
     \bottomrule[1pt]
     \end{tabular}
     }
  \end{table}

  \medskip\noindent\textbf{Hyperparameter Sensitivity Analysis.}\quad We adopt CSRA with ResNet-cut as a base model and apply MSL to evaluate its performance for various values of the trade-off hyperparameters $\alpha_{1}$, $\alpha_{2}$ and $\alpha_{3}$ on VOC2007. Table~\ref{Tab:abl_hyperparameters} shows the effect of each hyperparameter on MSL performance in terms of mAP, CR and CF1. Interestingly, the best performance is achieved when the trade-off hypeparameters $\alpha_{1}$ and $\alpha_{2}$ are weighted almost equally. Moreover, using label consistency with $\alpha_{3} = 0.5$ gives the best results. This suggests that the learned representations for partial inputs contribute to the improvement of the overall performance.

  \begin{table}[htp]
     \begin{center}
     \caption{Effect of hyperparameters on MSL performance using VOC2007.}
     \smallskip
     \label{Tab:abl_hyperparameters}
     \begin{tabular}{lccccc}
     \toprule[1pt]
     $\alpha_{1}$ & $\alpha_{2}$ & $\alpha_{3}$ & mAP & CR & CF1\\
     \midrule[1pt]
     1 & 1 & 1 & 93.6 & 87.7 & 88.2\\
     0.2 & 0.2 & 0.6 & 94.7 & 88.6 & 89.5\\
     0.3 & 0.3 & 0.4 & 95.0 & 89.0 & 89.9\\
     0.4 & 0.4 & 0.2 & 94.6 & 89.1 & 89.5\\
     \midrule[.8pt]
     0.3 & 0.2 & 0.5 & \textbf{96.1} & \textbf{92.4} & \textbf{91.6}\\
     \bottomrule[1pt]
     \end{tabular}
     \end{center}
  \end{table}


  \subsection{Robustness}
  We now examine the robustness of MSL against partial inputs and showcase its ability to predict non-masked objects.

\begin{figure}[!h]
     \centering
     \begin{subfigure}[b]{0.23\textwidth}
        \includegraphics[width=1\linewidth]{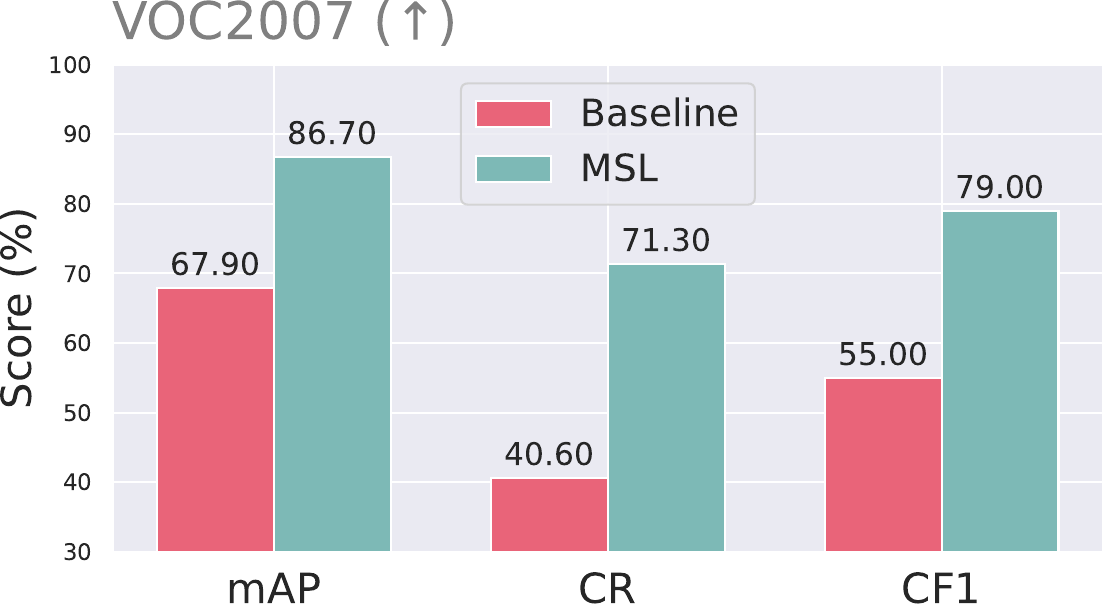}
        \caption{ResNet-cut on VOC2007.}
        \label{fig:Ng1}
     \end{subfigure}
     \hspace{0.1cm}
     \begin{subfigure}[b]{0.23\textwidth}
        \includegraphics[width=1\linewidth]{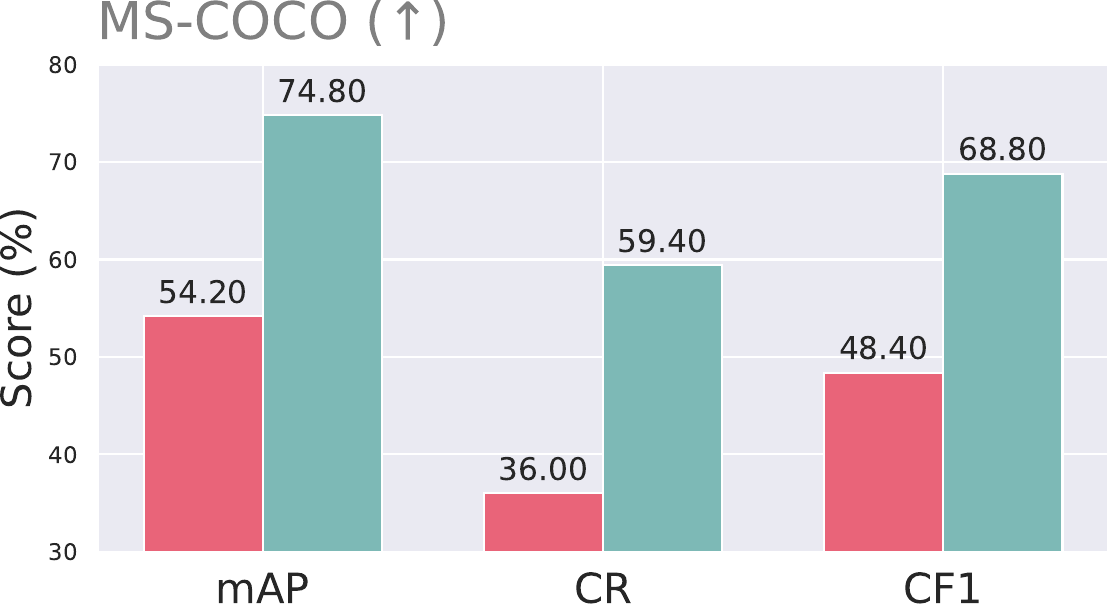}
        \caption{ResNet-cut on MS-COCO.}
        \label{fig:Ng2}
     \end{subfigure}
     \caption{\textbf{Performance comparisons when provided randomly masked images at test-time on VOC2007 and MS-COCO}. MSL is robust to heavily masked inputs, and hence occlusions.}
     \label{fig:robustness}
  \end{figure}

  \medskip\noindent\textbf{Quantitative Results.}\quad We evaluate MSL against partial inputs by deliberately masking the input images before making a prediction. In Figure~\ref{fig:robustness}(a), we show comparison results of MSL against CSRA with ResNet-cut on VOC2007. Using MSL, mAP is improved by 19.8\%, while CR and CF1 are improved by 30.7\% and 24\%, respectively. A similar trend is observed when comparing MSL against CSRA with ResNet-cut on MS-COCO, as depicted in Figure~\ref{fig:robustness}(b), achieving a mAP improvement of 20.6\%. This shows that MSL is robust to heavily masked inputs, and hence occlusions.

\begin{figure*}[!htb]
  \centering
  \includegraphics[scale=.195]{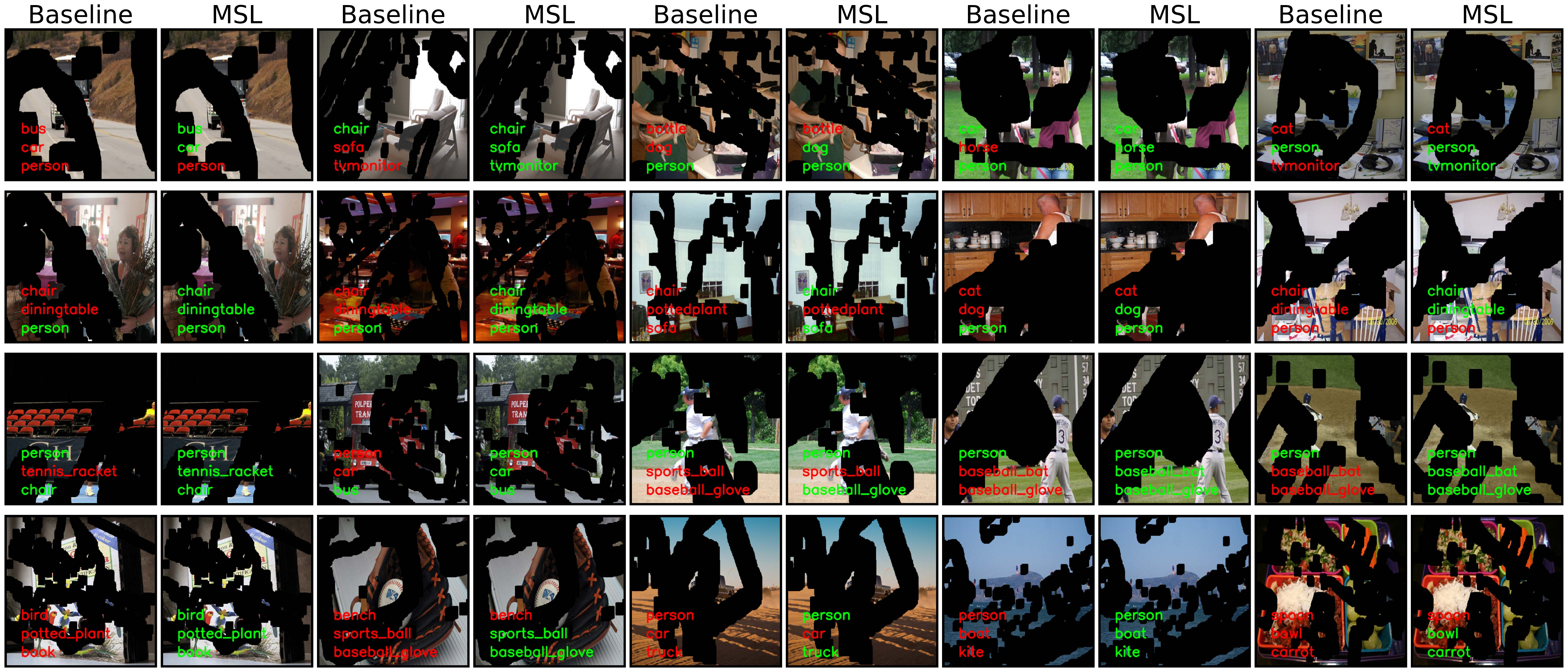}
  \caption{\textbf{Visual comparison of predictions made by MSL and baseline for masked input images on VOC2007 (first two rows) and MS-COCO (last two rows) datasets}. MSL is able recognize objects that are heavily masked and performs well in cases where the object is almost completely masked.}
  \label{Fig:qual_masked2}
  \end{figure*}

  \noindent\textbf{Qualitative Results.}\quad In Figure~\ref{Fig:qual_masked2}, we show visual comparisons of the top three predictions by our approach compared to the baseline when making predictions on masked inputs. We can see that the baseline fails to make predictions when the input image is masked. Even in cases where the object is slightly masked, the baseline fails to make a prediction. By comparison, our model is able to \textbf{recognize objects that are heavily masked} thanks to the masked branch. Also, there are cases where the \textbf{object is almost completely masked, but still our method is able to make a prediction}. This is largely attributed to label consistency, where the target label can be inferred from the other predicted labels.

\begin{figure}[!htb]
  \centering
  \includegraphics[scale=.11]{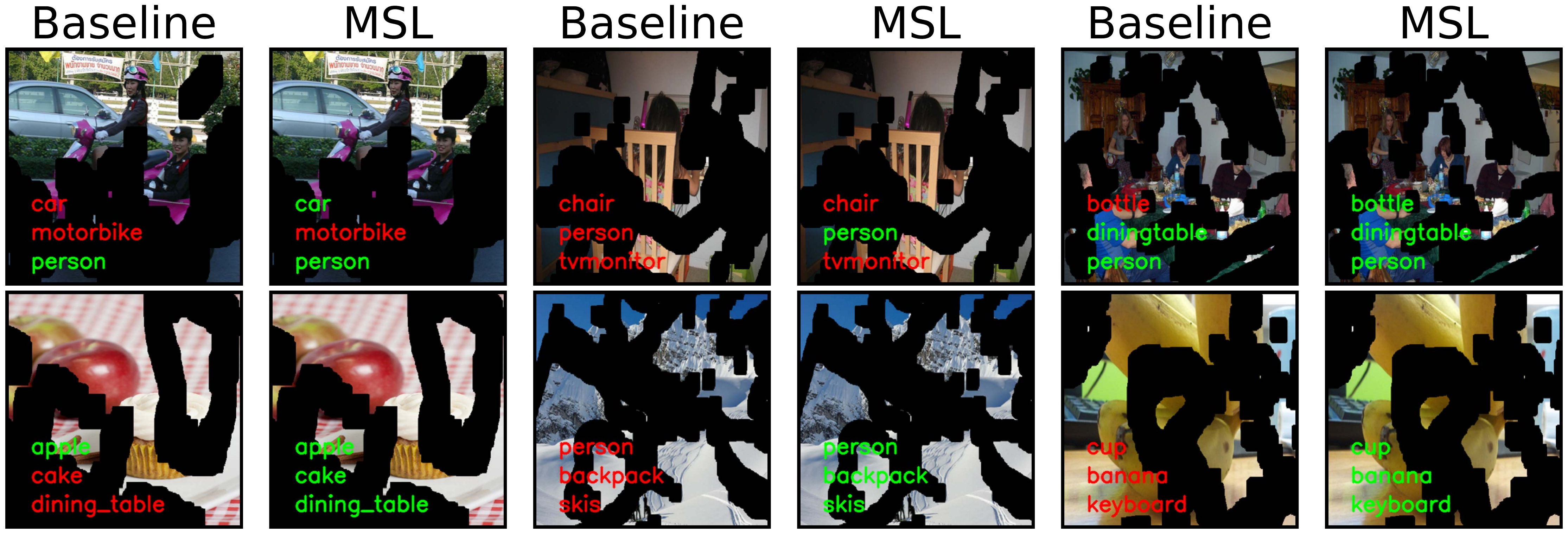}
  \caption{\textbf{Comparison of MSL and baseline on VOC2007 (first row) and MS-COCO (second row)}. Interestingly, MSL yields better prediction of non-masked objects.}
  \label{Fig:qual2}
  \end{figure}

\medskip\noindent\textbf{Non-Masked Objects.}\quad We also highlight an interesting property of MSL predictions in Figure~\ref{Fig:qual2}, which shows that our model \textbf{predicts non-masked objects} better than the baseline. We hypothesize that this is due in part to the initial features or cues that the model needs to focus on.

\smallskip\noindent\textbf{Comparison with random masking strategy.} While the Masked Autoencoder (MAE)~\cite{he2022masked} is a well-established masking strategy frequently employed in self-supervised learning, the key novelty of our MSL framework lies in the application of a masking strategy within the context of supervised learning. This novel utilization of masking during supervised learning sets our approach apart from existing methods. Moreover, MAE follows a two-step process: first, it undergoes pre-training for 800 epochs exclusively on images, and then it proceeds to fine-tune for an additional 50 epochs using both images and labels. In contrast, MSL requires only a single stage of training, lasting 60 epochs, utilizing both images and labels. To compare the performance of MSL and MAE, we present the results in Table~\ref{Tab:mae_compare}, which demonstrates the superiority of MSL over MAE in terms of mAP on both VOC2007 and MS-COCO datasets.

\begin{table}[!htb]
\setlength{\tabcolsep}{3.5pt}
\centering
\caption{\textbf{Performance comparison of MSL and MAE in mAP.}}
\smallskip
\label{Tab:mae_compare}
\begin{tabular}{l c c}
\toprule[1pt]
Masking & VOC2007 & MS-COCO\\
\midrule[1pt]
MAE~\cite{he2022masked} & 95.3 & 85.5 \\
MSL & \textbf{96.1} & \textbf{86.4}\\
\bottomrule[1pt]
\end{tabular}
\end{table}

\medskip\noindent\textbf{Comparison with CSRA variants.} In Table~\ref{Tab:csra}, we compare CSRA variants and MSL variants on VOC2007 and MS-COCO. As can be seen, MSL yields improved performance for both transformer and convolutional backbones.

\begin{table}[htp]
   \setlength{\tabcolsep}{3.5pt}
  \centering
  \caption{\textbf{Performance comparison of MSL and CSRA variants}.}
   \smallskip
   \label{Tab:csra}
   \resizebox{0.48\textwidth}{!}{%
   \begin{tabular}{l c c}
   \toprule[1pt]
   Method & VOC2007, mAP (\%)  & MS-COCO, mAP (\%)\\
   \midrule[.8pt]
   VIT-L16 & 92.1 & 75.6\\
   VIT-L16 w/ CSRA & 94.4 & 76.8\\
   VIT-L16 w/ MSL & \textbf{94.9} & \textbf{77.4}\\
   \midrule[.8pt]
   ResNet-Cut & 92.4 & 81.0\\
   ResNet-Cut w/ CSRA & 93.7 & 84.3\\
   ResNet-Cut w/ MSL & \textbf{94.4} & \textbf{85.5}\\
   \bottomrule[1pt]
   \end{tabular}
   }
\end{table}

\section{Conclusion}
In this paper, we presented a single-stage, model-agnostic learning paradigm using masking. The proposed paradigm, which is motivated by the intuition that occluded objects are \textit{partial inputs}, enables models to explicitly learn context-based representations and to model the label co-occurrence. We showed through extensive experiments that our method surpasses state-of-the-art models that heavily depend on multiple stages of training, high input resolution, the combination of multiple networks, large language models, complex data augmentation strategies, and additional data. We also demonstrated that MSL is robust to masked partial inputs for large and small objects, which is a strong indicator of its ability to handle challenging cases of small and occluded objects. Our method distinguishes itself from previous approaches due to its simple and straightforward training process, with the added benefit of incurring only a minor computational overhead compared to those methods. For future work, we aim to adapt the proposed framework to other computer vision tasks such as object detection.

\medskip\noindent\textbf{Acknowledgments.}\quad This work was supported in part by the Discovery Grants program of Natural Sciences and Engineering Research Council of Canada.

{\small
\bibliographystyle{ieee_fullname}
\bibliography{references}

\begin{thebibliography}{10}\itemsep=-1pt

\bibitem{ben2020asymmetric}
Emanuel Ben-Baruch, Tal Ridnik, Nadav Zamir, Asaf Noy, Itamar Friedman, Matan
  Protter, and Lihi Zelnik-Manor.
\newblock Asymmetric loss for multi-label classification.
\newblock In {\em Proc. IEEE International Conference on Computer Vision},
  2021.

\bibitem{bromley1993signature}
Jane Bromley, Isabelle Guyon, Yann LeCun, Eduard S{\"a}ckinger, and Roopak
  Shah.
\newblock Signature verification using a ``siamese'' time delay neural network.
\newblock In {\em Advances in Neural Information Processing Systems}, 1993.

\bibitem{LongChen:19}
Long Chen, Wujing Zhan, Wei Tian, Yuhang He, and Qin Zou.
\newblock Deep integration: A multi-label architecture for road scene
  recognition.
\newblock {\em IEEE Transactions on Image Processing}, 28:4883--4898, 2019.

\bibitem{chen2020knowledge}
Tianshui Chen, Liang Lin, Xiaolu Hui, Riquan Chen, and Hefeng Wu.
\newblock Knowledge-guided multi-label few-shot learning for general image
  recognition.
\newblock {\em IEEE Transactions on Pattern Analysis and Machine Intelligence},
  2020.

\bibitem{chen2018recurrent}
Tianshui Chen, Zhouxia Wang, Guanbin Li, and Liang Lin.
\newblock Recurrent attentional reinforcement learning for multi-label image
  recognition.
\newblock In {\em Proc. AAAI Conference on Artificial Intelligence}, volume~32,
  2018.

\bibitem{chen2019learning}
Tianshui Chen, Muxin Xu, Xiaolu Hui, Hefeng Wu, and Liang Lin.
\newblock Learning semantic-specific graph representation for multi-label image
  recognition.
\newblock In {\em Proc. IEEE International Conference on Computer Vision},
  pages 522--531, 2019.

\bibitem{chen2021learning}
Zhaomin Chen, Xiu-Shen Wei, Peng Wang, and Yanwen Guo.
\newblock Learning graph convolutional networks for multi-label recognition and
  applications.
\newblock {\em IEEE Transactions on Pattern Analysis and Machine Intelligence},
  2021.

\bibitem{chen2022sst}
Zhao-Min Chen, Quan Cui, Borui Zhao, Renjie Song, Xiaoqin Zhang, and Osamu
  Yoshie.
\newblock {SST}: Spatial and semantic transformers for multi-label image
  recognition.
\newblock {\em IEEE Transactions on Image Processing}, 31:2570--2583, 2022.

\bibitem{chen2019multi}
Zhao-Min Chen, Xiu-Shen Wei, Peng Wang, and Yanwen Guo.
\newblock Multi-label image recognition with graph convolutional networks.
\newblock In {\em Proc. IEEE International Conference on Computer Vision},
  pages 5177--5186, 2019.

\bibitem{devlin2018bert}
Jacob Devlin, Ming-Wei Chang, Kenton Lee, and Kristina Toutanova.
\newblock {BERT}: Pre-training of deep bidirectional transformers for language
  understanding.
\newblock In {\em Proc. NAACL-HLT}, pages 4171--4186, 2019.

\bibitem{dosovitskiy2020vit}
Alexey Dosovitskiy, Lucas Beyer, Alexander Kolesnikov, Dirk Weissenborn,
  Xiaohua Zhai, Thomas Unterthiner, Mostafa Dehghani, Matthias Minderer, Georg
  Heigold, Sylvain Gelly, Jakob Uszkoreit, and Neil Houlsby.
\newblock An image is worth 16x16 words: Transformers for image recognition at
  scale.
\newblock In {\em {International Conference on Learning Representations}},
  2021.

\bibitem{everingham2010pascal}
Mark Everingham, Luc Van~Gool, Christopher~KI Williams, John Winn, and Andrew
  Zisserman.
\newblock The {PASCAL} visual object classes {(VOC)} challenge.
\newblock {\em International Journal of Computer Vision}, 88(2):303--338, 2010.

\bibitem{gao2021learning}
Bin-Bin Gao and Hong-Yu Zhou.
\newblock Learning to discover multi-class attentional regions for multi-label
  image recognition.
\newblock {\em IEEE Transactions on Image Processing}, 30:5920--5932, 2021.

\bibitem{he2022masked}
Kaiming He, Xinlei Chen, Saining Xie, Yanghao Li, Piotr Doll{\'a}r, and Ross
  Girshick.
\newblock Masked autoencoders are scalable vision learners.
\newblock In {\em Proc. IEEE Conference on Computer Vision and Pattern
  Recognition}, 2022.

\bibitem{he2016deep}
Kaiming He, Xiangyu Zhang, Shaoqing Ren, and Jian Sun.
\newblock Deep residual learning for image recognition.
\newblock In {\em Proc. IEEE Conference on Computer Vision and Pattern
  Recognition}, pages 770--778, 2016.

\bibitem{lanchantin2021general}
Jack Lanchantin, Tianlu Wang, Vicente Ordonez, and Yanjun Qi.
\newblock General multi-label image classification with transformers.
\newblock In {\em Proc. IEEE Conference on Computer Vision and Pattern
  Recognition}, pages 16478--16488, 2021.

\bibitem{li2020bi}
Peng Li, Peng Chen, Yonghong Xie, and Dezheng Zhang.
\newblock Bi-modal learning with channel-wise attention for multi-label image
  classification.
\newblock {\em IEEE Access}, 8:9965--9977, 2020.

\bibitem{li2016human}
Yining Li, Chen Huang, Chen~Change Loy, and Xiaoou Tang.
\newblock Human attribute recognition by deep hierarchical contexts.
\newblock In {\em Proc. European Conference on Computer Vision}, pages
  684--700, 2016.

\bibitem{lin2014microsoft}
Tsung-Yi Lin, Michael Maire, Serge Belongie, James Hays, Pietro Perona, Deva
  Ramanan, Piotr Doll{\'a}r, and C~Lawrence Zitnick.
\newblock Microsoft {COCO}: Common objects in context.
\newblock In {\em Proc. European Conference on Computer Vision}, pages
  740--755. Springer, 2014.

\bibitem{liu2018image}
Guilin Liu, Fitsum~A Reda, Kevin~J Shih, Ting-Chun Wang, Andrew Tao, and Bryan
  Catanzaro.
\newblock Image inpainting for irregular holes using partial convolutions.
\newblock In {\em Proc. European Conference on Computer Vision}, pages 85--100,
  2018.

\bibitem{liu2022causality}
Ruyang Liu, Jingjia Huang, Thomas~H Li, and Ge Li.
\newblock Causality compensated attention for contextual biased visual
  recognition.
\newblock In {\em International Conference on Learning Representations}, 2023.

\bibitem{liu2021query2label}
Shilong Liu, Lei Zhang, Xiao Yang, Hang Su, and Jun Zhu.
\newblock {Query2label}: A simple transformer way to multi-label
  classification.
\newblock {\em arXiv preprint arXiv:2107.10834}, 2021.

\bibitem{liu2018multi}
Yongcheng Liu, Lu Sheng, Jing Shao, Junjie Yan, Shiming Xiang, and Chunhong
  Pan.
\newblock Multi-label image classification via knowledge distillation from
  weakly-supervised detection.
\newblock In {\em Proc. ACM International Conference on Multimedia}, pages
  700--708, 2018.

\bibitem{suvorov2022resolution}
Roman Suvorov, Elizaveta Logacheva, Anton Mashikhin, Anastasia Remizova,
  Arsenii Ashukha, Aleksei Silvestrov, Naejin Kong, Harshith Goka, Kiwoong
  Park, and Victor Lempitsky.
\newblock Resolution-robust large mask inpainting with fourier convolutions.
\newblock In {\em Proc. IEEE Winter Conference on Applications of Computer
  Vision}, pages 2149--2159, 2022.

\bibitem{wang2017multi}
Zhouxia Wang, Tianshui Chen, Guanbin Li, Ruijia Xu, and Liang Lin.
\newblock Multi-label image recognition by recurrently discovering attentional
  regions.
\newblock In {\em Proc. IEEE International Conference on Computer Vision},
  pages 464--472, 2017.

\bibitem{yang2016exploit}
Hao Yang, Joey Tianyi~Zhou, Yu Zhang, Bin-Bin Gao, Jianxin Wu, and Jianfei Cai.
\newblock Exploit bounding box annotations for multi-label object recognition.
\newblock In {\em Proc. IEEE Conference on Computer Vision and Pattern
  Recognition}, pages 280--288, 2016.

\bibitem{yazici2020orderless}
Vacit~Oguz Yazici, Abel Gonzalez-Garcia, Arnau Ramisa, Bartlomiej Twardowski,
  and Joost van~de Weijer.
\newblock Orderless recurrent models for multi-label classification.
\newblock In {\em Proc. IEEE Conference on Computer Vision and Pattern
  Recognition}, pages 13440--13449, 2020.

\bibitem{ye2020attention}
Jin Ye, Junjun He, Xiaojiang Peng, Wenhao Wu, and Yu Qiao.
\newblock Attention-driven dynamic graph convolutional network for multi-label
  image recognition.
\newblock In {\em Proc. European Conference on Computer Vision}, pages
  649--665. Springer, 2020.

\bibitem{you2020cross}
Renchun You, Zhiyao Guo, Lei Cui, Xiang Long, Yingze Bao, and Shilei Wen.
\newblock Cross-modality attention with semantic graph embedding for
  multi-label classification.
\newblock In {\em Proc. AAAI Conference on Artificial Intelligence}, volume~34,
  pages 12709--12716, 2020.

\bibitem{yun2019cutmix}
Sangdoo Yun, Dongyoon Han, Seong~Joon Oh, Sanghyuk Chun, Junsuk Choe, and
  Youngjoon Yoo.
\newblock {CutMix}: Regularization strategy to train strong classifiers with
  localizable features.
\newblock In {\em Proc. IEEE International Conference on Computer Vision},
  pages 6023--6032, 2019.

\bibitem{zhao2021transformer}
Jiawei Zhao, Ke Yan, Yifan Zhao, Xiaowei Guo, Feiyue Huang, and Jia Li.
\newblock Transformer-based dual relation graph for multi-label image
  recognition.
\newblock In {\em Proc. IEEE International Conference on Computer Vision},
  pages 163--172, 2021.

\bibitem{zhu2021residual}
Ke Zhu and Jianxin Wu.
\newblock Residual attention: A simple but effective method for multi-label
  recognition.
\newblock In {\em Proc. IEEE International Conference on Computer Vision},
  pages 184--193, 2021.

\bibitem{Zunair2022MSL}
Hasib Zunair and A. {Ben Hamza}.
\newblock Masked supervised learning for semantic segmentation.
\newblock In {\em Proc. British Machine Vision Conference}, 2022.

\end{thebibliography}
}

\clearpage
\setcounter{page}{1}
\section*{\LARGE{--- Supplementary Material ---}}
\section{Implementation details}

\noindent\textbf{Data preprocessing.}\quad Images and masks are resized to $448 \times 448$ and normalized to have values in $[0, 1]$. For ViT~\cite{dosovitskiy2020vit} based models, the input resolution is set to $224 \times 224$ to leverage ImageNet-21k and ImageNet pretrained weights. We use simple data augmentation techniques such as random flip and random resize crop. Unlike previous works~\cite{ben2020asymmetric,zhu2021residual}, we do not employ complex data augmentation strategies such as CutMix, GPU Augmentations, or RandAugment.

\medskip\noindent\textbf{Architecture.}\quad We apply MSL on two CSRA~\cite{zhu2021residual} based backbones, a convolutional backbone ResNet-cut which is a ResNet-101 pretrained on ImageNet with CutMix~\cite{yun2019cutmix} augmentation strategy. It is worth mentioning that we do not use CutMix~\cite{yun2019cutmix} augmentation strategy when applying MSL, to demonstrate its effectiveness. Note that here CutMix is for the pretrained model and not during fine-tuning on VOC2007 and MS-COCO datasets. To demonstrate the generality of MSL, we use a transformer backbone ViT-L16~\cite{dosovitskiy2020vit} pretrained on ImageNet-21k and fine-tuned on ImageNet with the $224 \times 224$ resolution. We drop class tokens and use the final output embeddings as features, and we also interpolate positional embeddings when the models are fine-tuned on the higher resolution datasets. We refer to these MSL variants as MSL-C and MSL-V, where C and V denote convolutional and vision transformer, respectively.

\medskip\noindent\textbf{Model Training.}\quad MSL models are trained in a single stage, requiring a training set comprised of images and labels. We use the SGD optimizer to minimize the loss function. Following previous work~\cite{zhu2021residual}, we apply simple data augmentation such as random flip and random resize crop. For training both the baseline and MSL models, we set the learning rate, momentum and weight decay to 0.01, 0.9 and 0.0001, respectively. The models are trained for 60 epochs with a batch size of 6, and the best weights according to the mAP score on the test set are recorded. We follow CSRA~\cite{zhu2021residual} models and set $H = 1$, $\lambda = 0.1$ for VOC2007, and $H = 6$, $\lambda = 0.4$ for MS-COCO.

\medskip\noindent\textbf{Model Testing.}\quad After training, given an image as input, the model simply makes a prediction by assigning multiple label(s) among the defined classes.

\medskip\noindent\textbf{Hardware and software details.}\quad Our experiments were conducted on a Linux workstation running 4.8Hz and 64GB RAM, equipped with a single NVIDIA RTX 3080Ti GPU packed with 12GB of memory. All algorithms are implemented in Python using PyTorch.

\section{Additional Results}
In this section, we provide additional experimental results on VOC2007, MS-COCO and WIDER-Attribute datasets, showing the effectiveness of MSL in recognizing small and occluded objects.

\medskip\noindent\textbf{Runtime Analysis.}\quad MSL incurs a minor computational overhead compared to traditional supervised learning. This is primarily due to the masking operation and the computation of predictions on the masked images. It is important to mention that this extra cost is only present during the training phase, and during inference, there is no masking involved. Instead, predictions are directly computed on the original input images. When compared to previous approaches, our method stands out for its simplicity and ease of training. Unlike other methods, MSL does not require multiple stages of training, the combination of multiple learnable networks, the utilization of large language models, high input resolution, complex data augmentation strategies, or the inclusion of additional data.

\begin{figure*}[htp]
   \centering
   \includegraphics[scale=.22]{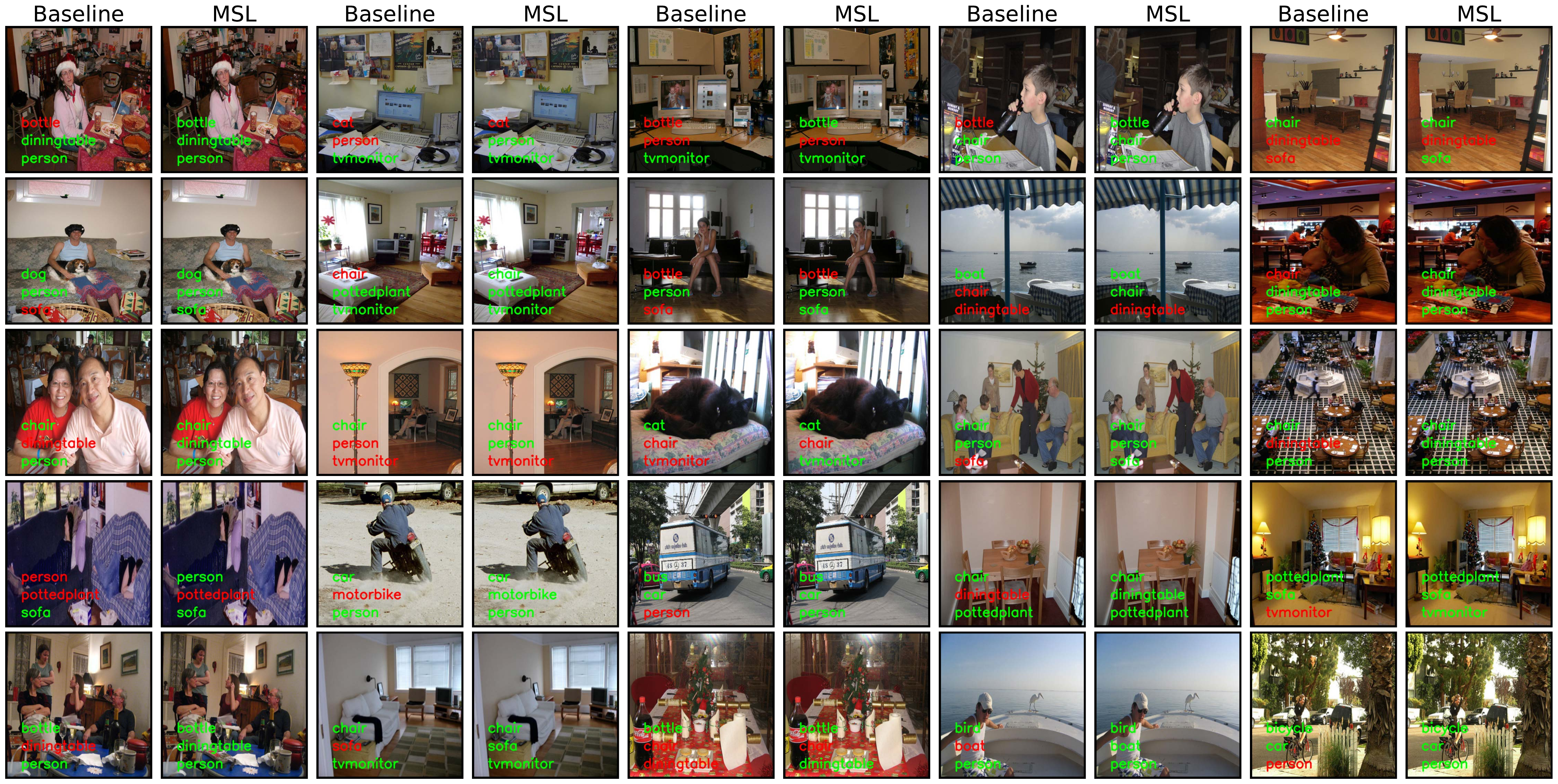}
   \caption{\textbf{Visual comparison of predictions of MSL and baseline on the VOC2007}. MSL is able to accurately predict small objects as well as objects under heavy occlusions.}
   \label{Fig:voc}
\end{figure*}

\begin{figure*}[htp]
   \centering
   \includegraphics[scale=.27]{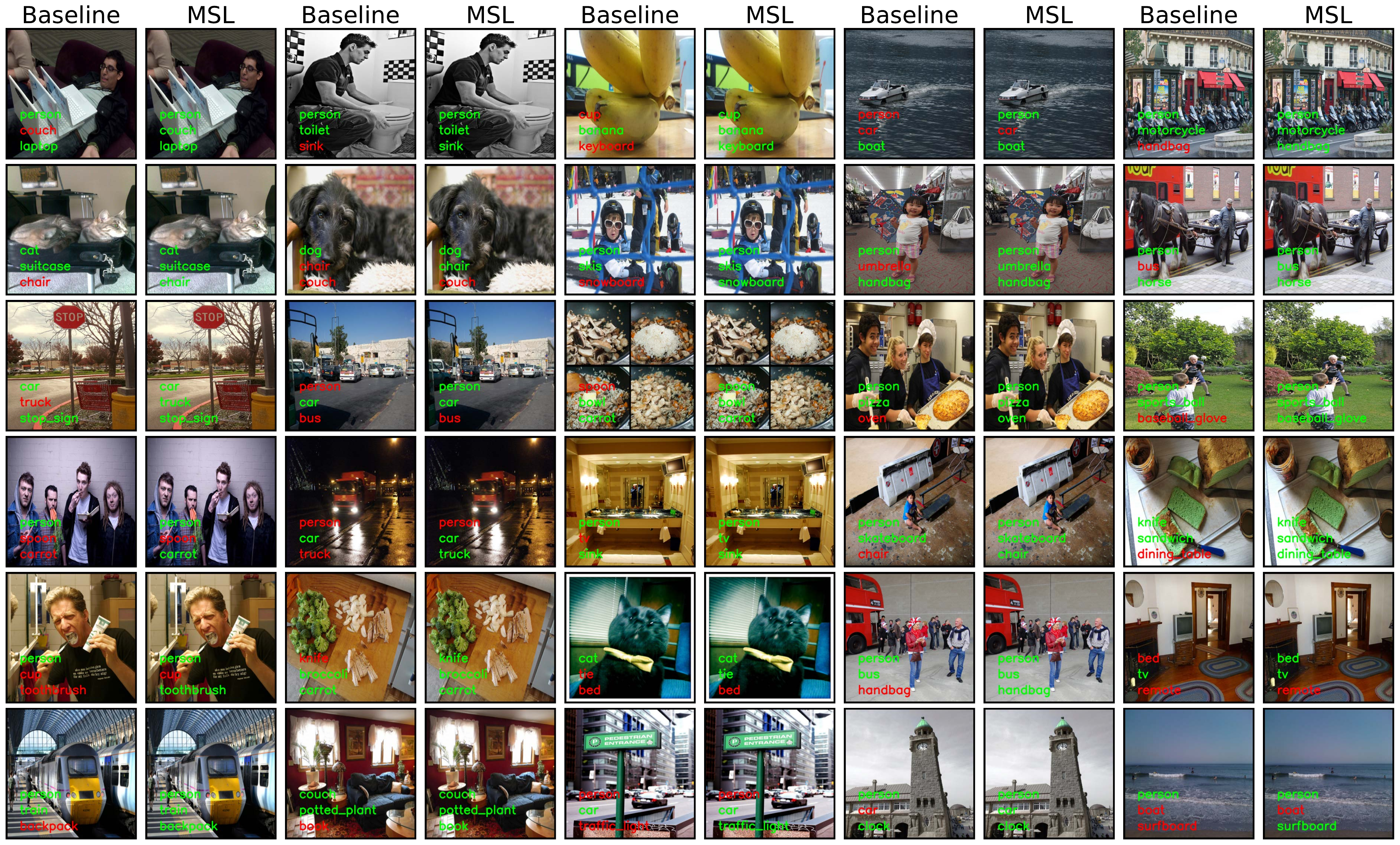}
   \caption{\textbf{Visual comparison of predictions of MSL and baseline on the MS-COCO test set}. MSL is able to accurately predict small objects as well as objects under heavy occlusions.}
   \label{Fig:coco}
\end{figure*}

\medskip\noindent\textbf{Discussion on MLIR for small objects.} Upon analyzing recent MLIR methods, we noticed that MCAR~\cite{gao2021learning} stands out as the only method that explicitly tackles the problem of small-sized and occluded objects. Comparatively, our MSL model achieves higher scores in terms of mean Average Precision (mAP), with values of 96.1\% and 86.4\% on the VOC2007 and MS-COCO datasets, respectively. On the other hand, MCAR's performance falls slightly behind, scoring 94.8\% and 84.5\% on the same datasets. Note that MCAR employs an input resolution of $576 \times 576$, while MSL operates at a resolution of $448 \times 448$. MSL explicitly addresses the problem of small and occluded objects through the Masked Branch since that task of the branch is to recognize masked objects, which are partial inputs. We further illustrate the effectiveness of MSL in handling small objects and heavily occluded objects through visual examples presented in Figures~\ref{Fig:voc} and \ref{Fig:coco}. These examples demonstrate MSL's ability to accurately predict such challenging instances.

\medskip\noindent\textbf{MSL is model-agnostic.} In Tables~\ref{Tab:msl_recent} and \ref{Tab:abl_modules}, we show recent state-of-the-art methods, as well as convolutional and transformer backbones, all of which were trained using MSL. As can be seen, MSL consistently improves performance of various methods, demonstrating that MSL is model-agnostic.

\begin{table}[htp]
   \setlength{\tabcolsep}{3.5pt}
	\centering
	\caption{\textbf{Comparison of recent architectures trained using MSL}. MSL is a versatile approach that enhances the performance of different methods. Note that MCAR models are trained using $576 \times 576$ input resolution, while the others utilized a resolution of $448 \times 448$. MSL improves performance of recent MLIR methods.}
   \smallskip
   \label{Tab:msl_recent}
   \resizebox{0.35\textwidth}{!}{%
   \begin{tabular}{l c}
   \toprule[1pt]
   Method & VOC2007, mAP (\%) \\
   \midrule[.8pt]
   MCAR~\cite{gao2021learning} & 94.8 \\
   MCAR~\cite{gao2021learning} w/ MSL & \textbf{95.6} \\
   SST~\cite{chen2022sst} & 94.5 \\
   SST~\cite{chen2022sst} w/ MSL & \textbf{95.8} \\
   \bottomrule[1pt]
   \end{tabular}
   }
\end{table}

\begin{table}[htp]
   \begin{center}
   \caption{\textbf{Comparison of different architectures trained using MSL on VOC2007}. MSL improves performance of both convolutional and transformer baselines in terms of mAP and other metrics.}
    \smallskip
   \label{Tab:abl_modules}
   \begin{tabular}{l c c c c}
   \toprule[1.1pt]
   Method & mAP & CR & CF1\\
   \midrule[1pt]
   ViT~\cite{dosovitskiy2020vit} & 94.4 & 86.9 & 89.6 \\
   + MSL & \textbf{95.0} & \textbf{84.8} & \textbf{89.5} \\
   \midrule[1pt]
   ResNet-cut~\cite{zhu2021residual} & 93.7 & 87.5 & 88.3 \\
    + MSL & \textbf{96.1} & \textbf{92.4} & \textbf{91.6}\\
   \bottomrule[1.1pt]
   \end{tabular}
   \end{center}
\end{table}

\medskip\noindent\textbf{WIDER-Attribute dataset results.} Table~\ref{Tab:wider} shows that MSL outperforms strong baselines on the WIDER-Attribute dataset~\cite{li2016human}.

\begin{table}[htp]
\begin{center}
\caption{\textbf{Performance comparison of MSL and baselines on WIDER-Attribute dataset}. MSL outperforms all baselines. $\dagger$ indicates our reproduced result. Other results are taken from~\cite{zhu2021residual}.}
\smallskip
\label{Tab:wider}
\begin{tabular}{l c c c c}
\toprule[1.1pt]
Method & mAP & CF1 & OF1\\
\midrule[.8pt]
DHC & 81.3 & - & - \\
VA & 82.9 & - & - \\
SRN & 86.2 & 75.9 & 81.3 \\
VAC & 87.5 & 77.6 & 82.4 \\
VIT-B16 & 86.3 & 75.9 & 81.5 \\
VIT-L16 & 87.7 & 78.1 & 82.8 \\
\midrule[.8pt]
VIT-L16 + CSRA$\dagger$ & 89.6 & 80.4 & 84.9 \\
VIT-L16 + MSL & \textbf{90.6} & \textbf{80.5} & \textbf{85.3}\\
\bottomrule[1.1pt]
\end{tabular}
\end{center}
\end{table}

\medskip\noindent\textbf{Comparison with CSRA variants.} In Table~\ref{Tab:csra}, a comparison is made between CSRA variants and MSL variants on VOC2007 and MS-COCO. Specifically, we train CSRA and MSL with two pretrained backbones, namely ViT-L16 and ResNet with CutMix. Note that in the main body of the paper, we use CSRA-based backbones in MSL with MSL-C and MSL-V notations. Here, we test CSRA and MSL independently to highlight the contributions of MSL. We find that MSL improves performance for both transformer and convolutional backbones on both datasets. For fair comparison, we run CSRA variants on our working environment and conduct all experiments with a batch size of 6, whereas the CSRA results reported in the paper~\cite{zhu2021residual} use a batch size of 64. Hence, the results we report here do not exactly match those in~\cite{zhu2021residual}. To analyze the effect of batch size on the performance of CSRA and MSL, we conduct a small experiment on VOC2007 by varying the batch size from 4 to 12, which maximizes our GPU usage, and we found that both CSRA and MSL improve in terms of performance. Therefore, we argue that the performance of MSL could be further improved using a higher batch size.

\begin{table}[htp]
   \setlength{\tabcolsep}{3.5pt}
  \centering
  \caption{\textbf{Performance comparison of MSL and CSRA variants in terms of mAP (\%) on VOC2007 and MS-COCO}. MSL outperforms CSRA variants on both datasets.}
   \smallskip
   \label{Tab:csra}
   \resizebox{0.45\textwidth}{!}{%
   \begin{tabular}{l c c}
   \toprule[1pt]
   Method & VOC2007, mAP (\%)  & MS-COCO, mAP (\%)\\
   \midrule[.8pt]
   VIT-L16 & 92.1 & 75.6\\
   VIT-L16 w/ CSRA & 94.4 & 76.8\\
   VIT-L16 w/ MSL & \textbf{94.9} & \textbf{77.4}\\
   \midrule[.8pt]
   ResNet-Cut & 92.4 & 81.0\\
   ResNet-Cut w/ CSRA & 93.7 & 84.3\\
   ResNet-Cut w/ MSL & \textbf{94.4} & \textbf{85.5}\\
   \bottomrule[1pt]
   \end{tabular}
   }
\end{table}

\medskip\noindent\textbf{Analysis of masking in MSL.} In Table~\ref{Tab:masking_ext}, we report the impact of low and high masking on the performance of MSL-C and MSL-V. As can be seen, better results are achieved with high masking on different backbones tested on both VOC2007 and MS-COCO. High masking enables the network to learn better context when training using MSL. Low masking, on the other hand, does not result in significant performance improvements, partly due to learning redundant features. In other words, low masking does not significantly change the original image. Hence, learning very similar features does not help to learn useful representations.

\begin{table}[htp]
   \begin{center}
   \caption{\textbf{Ablation analysis in mAP (\%) of high- and low-masked pixels during MSL training on VOC2007 and MS-COCO.} MSL with high-masked pixels yields better performance.}
   \smallskip
   \label{Tab:masking_ext}
   \begin{tabular}{l c c c}
   \toprule[1pt]
   Method & Masking & VOC2007& MS-COCO\\
  \midrule[.8pt]
   MSL-V & Low & 94.6 & 77.8\\
   MSL-V & High & \textbf{95.0} & \textbf{79.0}\\
   \midrule[.8pt]
   MSL-C & Low & 95.0 & 85.1\\
   MSL-C & High & \textbf{96.1} & \textbf{86.4}\\
   \bottomrule[1pt]
   \end{tabular}
   \end{center}
\end{table}

\newpage

\begin{table*}[htp]
   \begin{center}
   \caption{\textbf{Performance comparisons of MSL and CSRA ResNet-cut~\cite{zhu2021residual} as baseline in terms of mAP (\%) and other metrics when provided randomly masked images at test time on VOC2007 and MS-COCO datasets}. Boldface numbers indicate the best performance. MSL is more robust to partial inputs.}
   \smallskip
   \label{Tab:supp_rob}
   \begin{tabular}{l c c c c c c c}
   \toprule[1.1pt]
   Method & mAP & CP & CR & CF1 & OP & OR & OF1\\
   \midrule[1pt]
   Baseline (VOC2007) & 67.9 & 85.5 & 40.6 & 55.1 & 75.2 & 48.7 & 59.1\\
   + w/o MSL & \textbf{86.7} & \textbf{88.6} & \textbf{71.2} & \textbf{79.0} & \textbf{91.8} & \textbf{73.7} & \textbf{81.8}\\
   \midrule[1pt]
   Baseline (MS-COCO) & 54.2 & 73.6 & 36.0 & 48.4 & 69.1 & 44.8 & 54.4\\
   + w/o MSL & \textbf{74.8} & \textbf{81.9} & \textbf{59.3} & \textbf{68.8} & \textbf{84.5} & \textbf{64.0} & \textbf{72.9}\\
   \bottomrule[1.1pt]
   \end{tabular}
   \end{center}
\end{table*}

\begin{figure*}[htp]
   \centering
   \includegraphics[scale=.27]{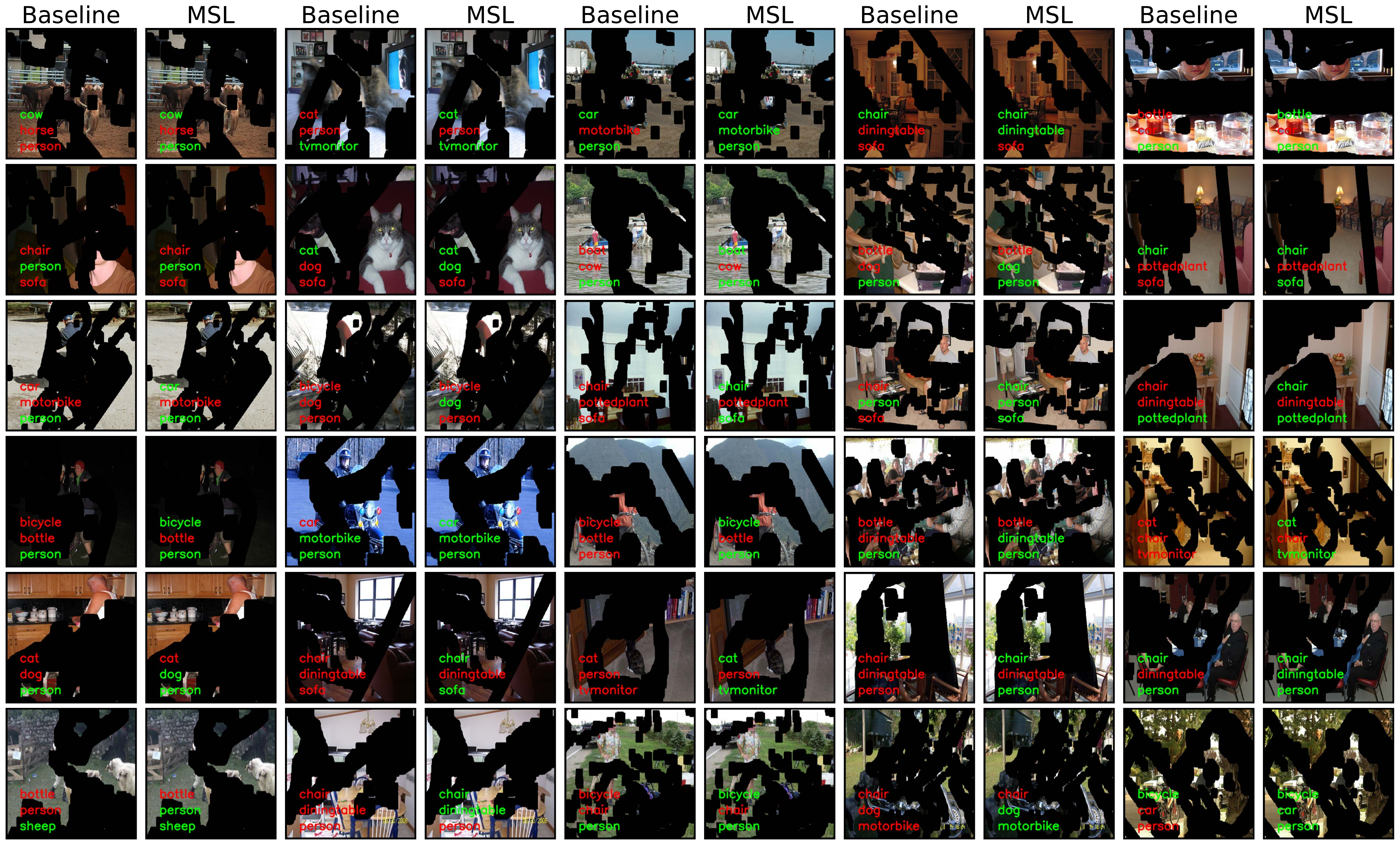}
   \caption{\textbf{Visual comparison of predictions of MaskSup and baseline on the VOC2007 test set when provided with \textbf{masked regions} as input}. MSL is able to recognize objects that are heavily masked and even recognize objects that are almost completely masked.}
   \label{Fig:voc_masked}
\end{figure*}

\begin{figure*}[htp]
   \centering
   \includegraphics[scale=.27]{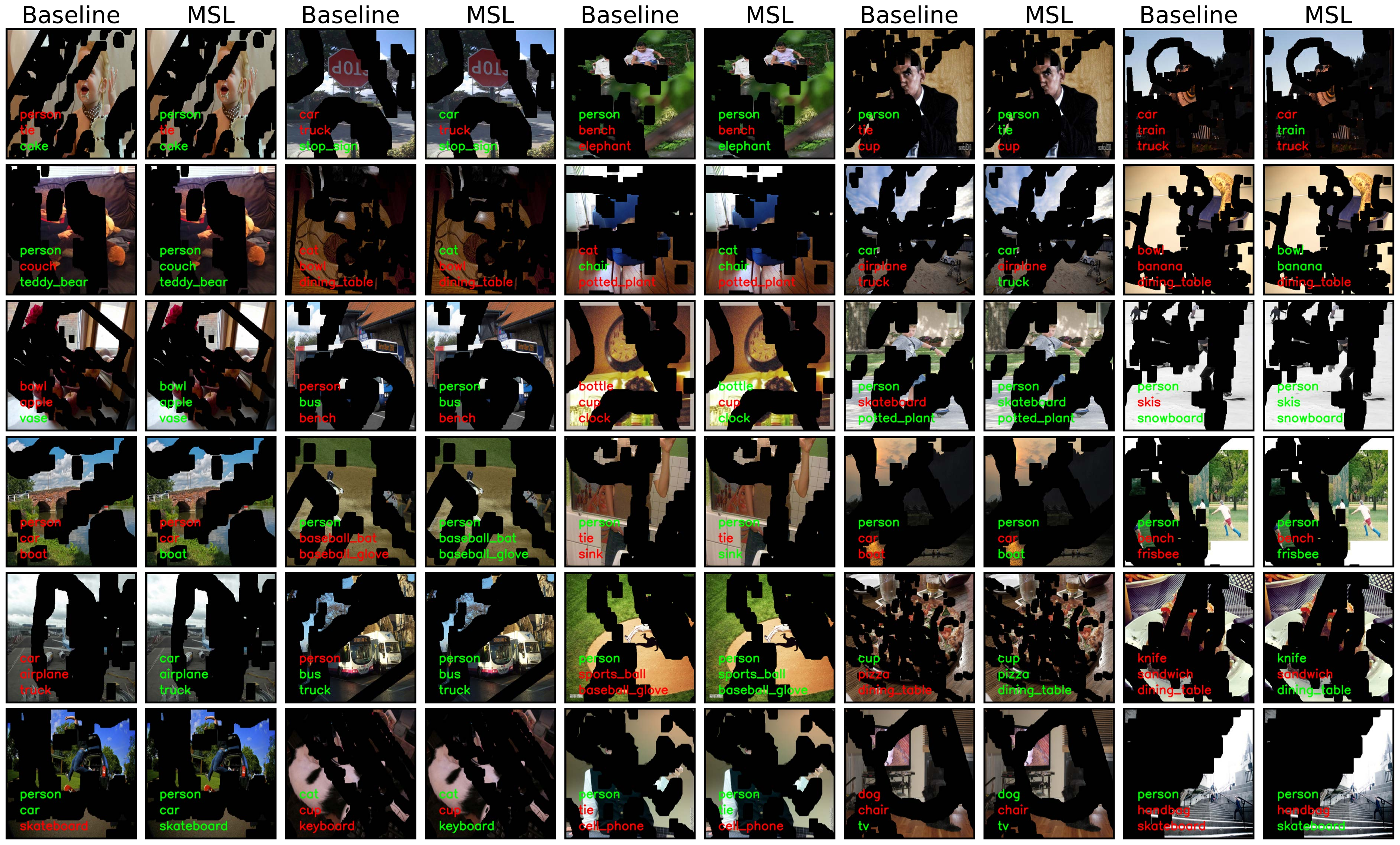}
   \caption{\textbf{Visual comparison of predictions of MaskSup and the strongest baseline on the MS-COCO test set when provided with \textbf{masked regions} as input}. MSL is able to recognize objects that are heavily masked and even recognize objects that are almost completely masked..}
   \label{Fig:coco_masked}
\end{figure*}

\begin{figure*}[htp]
   \centering
   \includegraphics[scale=.20]{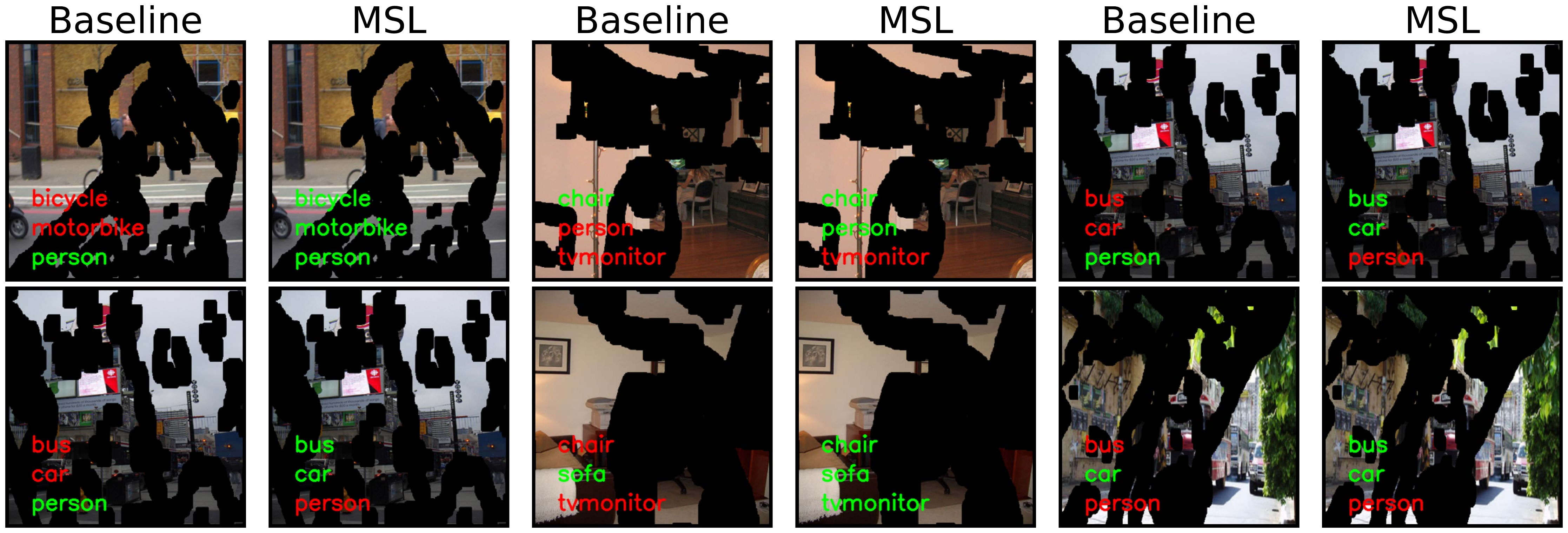}
   \caption{\textbf{Comparison of MSL and the strongest baseline on the VOC2007 and MS-COCO test sets in first and second rows}. It is worth noting that MSL is able to predict non-masked objects that the baseline model often fails to detect.}
   \label{Fig:voc_non_masked}
\end{figure*}

\begin{figure*}[htp]
   \centering
   \includegraphics[scale=.45]{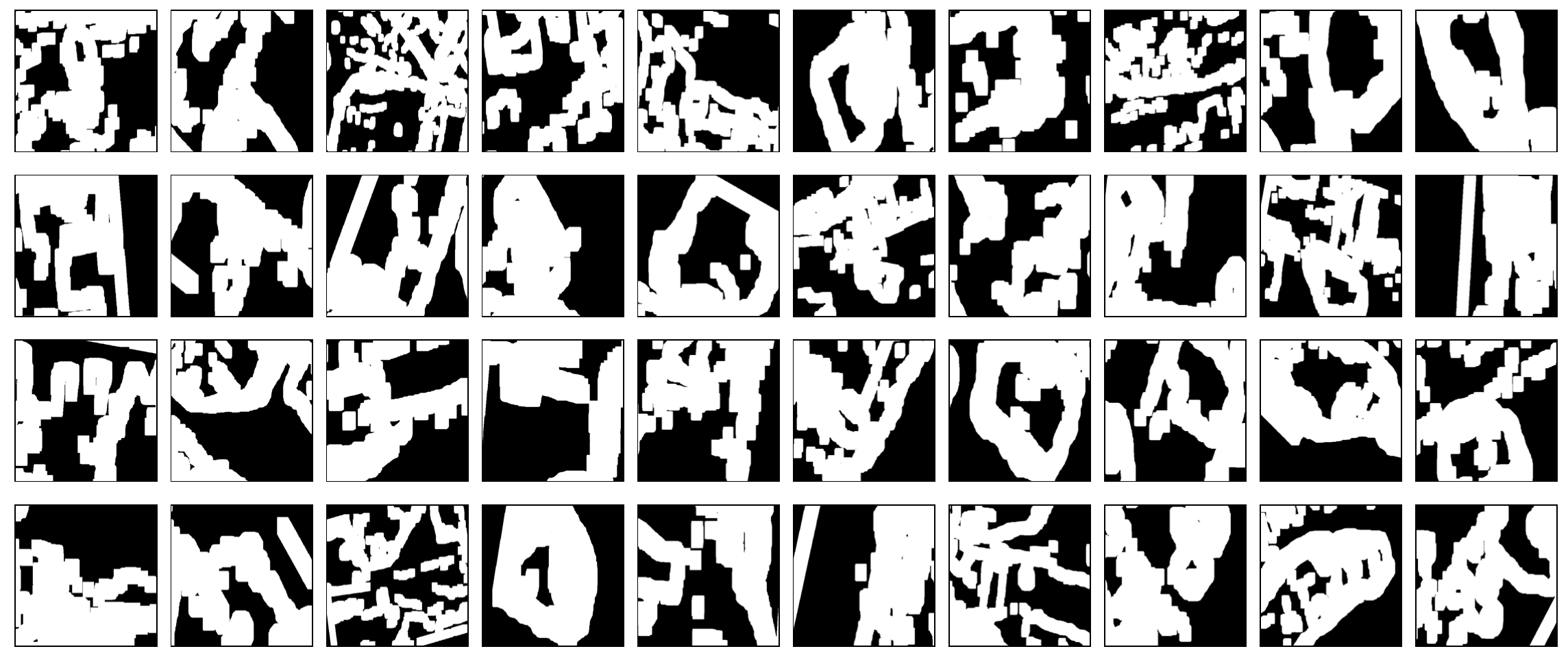}
   \caption{\textbf{Visual of masks during training in MSL}. Some masks cover more than 50\% of the image. Images are from Irregular Masks Dataset~\cite{liu2018image} after applying
   binary thresholding.}
   \label{Fig:mask}
\end{figure*}

\end{document}